\documentclass[runningheads]{llncs}
\usepackage{graphicx}
\begin{document}
\title{Social Media as a Sensor: Analyzing Twitter Data for Breast Cancer Medication Effects Using Natural Language Processing}
\titlerunning{Breast cancer medication side effect discovery using NLP}
%
\author{Seibi Kobara\textsuperscript{†}\inst{1}\orcidID{0000-0001-8134-7652} \and
Alireza Rafiei\textsuperscript{†}\inst{1}\orcidID{0000-0001-9747-4023} \and
Masoud Nateghi\inst{1}\orcidID{0009-0002-4263-2092} \and
Selen Bozkurt\inst{1}\orcidID{0000-0003-1234-2158} \and
Rishikesan Kamaleswaran \inst{1}\orcidID{0000-0001-8366-4811} \and
Abeed Sarker \inst{1}\orcidID{0000-0001-7358-544X}
\newline
†: Equally contributed to this paper}

\authorrunning{S. Kobara et al.}
%
\institute{Emory University, 201 Dowman Dr, Atlanta, GA 30322, USA}

\maketitle              
%

\begin{abstract} 
Breast cancer is a significant public health concern and is the leading cause of cancer-related deaths among women. Despite advances in breast cancer treatments, medication non-adherence remains a major problem. As electronic health records do not typically capture patient-reported outcomes that may reveal information about medication-related experiences, social media presents an attractive resource for enhancing our understanding of the patients’ treatment experiences. In this paper, we developed natural language processing (NLP) based methodologies to study information posted by an automatically curated breast cancer cohort from social media. We employed a transformer-based classifier to identify breast cancer patients/survivors on X (Twitter) based on their self-reported information, and we collected longitudinal data from their profiles. We then designed a multi-layer rule-based model to develop a breast cancer therapy-associated side effect lexicon and detect patterns of medication usage and associated side effects among breast cancer patients. 1,454,637 posts were available from 583,962 unique users, of which 62,042 were detected as breast cancer members using our transformer-based model. 198 cohort members mentioned breast cancer medications with tamoxifen as the most common. Our side effect lexicon identified well-known side effects of hormone and chemotherapy. Furthermore, it discovered a subject feeling towards cancer and medications, which may suggest a pre-clinical phase of side effects or emotional distress. This analysis highlighted not only the utility of NLP techniques in unstructured social media data to identify self-reported breast cancer posts, medication usage patterns, and treatment side effects but also the richness of social data on such clinical questions.


\end{abstract}

\section{Introduction}
Breast cancer, the most prevalent cancer among women, represents a significant public health concern. Accounting for about 30\% of all new female cancer cases annually, it stands as the second leading cause of cancer death in women, following lung cancer \cite{pmid35143040,miller2021cancer}. Despite the grim statistics, there has been a consistent decrease in breast cancer mortality rates since 1989, with an overall decline of 43\% through 2020 \cite{pmid36190501}. This notable progress is attributed to earlier diagnosis, increased awareness, and advancements in treatments \cite{pmid16251534,pmid25255803,pmid29963498}. However, the pace of this decline has shown signs of slowing in recent years \cite{pmid36190501}, emphasizing the need for continued research and innovation in breast cancer care.

\par Despite the advances in breast cancer treatments, including endocrine therapy, which have led to declining death rates, as many as half to two-thirds of these breast cancer patients discontinue endocrine treatment within the first three years, increasing the risk of recurrence, hospitalization, and even death \cite{waterhouse1993adherence,mccowan2008cohort}.
Treatment non-adherence and discontinuation are often due to medication-related physical and mental side effects. 
Treatment-related side effects, or other subtle factors leading to non-adherence, are not detectable by laboratory diagnostic tests, but can be gathered through patient communications. Information gleaned from patient communications (i.e., patient-reported outcomes (PROs)) are sometimes captured as free text in clinical narratives or through patient surveys. Both mechanisms of documenting information are labor-intensive and subject to biases. Furthermore, electronic health records (EHRs) have been found to under-document PROs (e.g., only 8\% of a sample were found to contain PROs in a study) \cite{mccowan2008cohort}. 
Thus, studies based solely on EHRs or other traditional instruments can only capture limited and instrument-specific clinical information, often relating to cancer control endpoints.

PROs or patient experiences are crucial in understanding the overall impact of breast cancer therapies and guiding future treatment strategies. One potential source of such information is social media, where patients are known to discuss their experiences with their peers. The potential of social media, specifically X (formerly known as Twitter), in this context is particularly compelling, because of its vast and diverse user base, and its ability to serve as a real-time global sensor for public sentiment and personal experiences. By tapping into the rich, unstructured data of social media, the trends and patient experiences that might remain hidden in clinical settings can be uncovered, tailoring for more patient-centered healthcare practices. However, obtaining information from real breast cancer survivors requires the establishment of a social media based cohort, and then analyzing data posted by this cohort. Natural language processing (NLP) and machine learning methods provide potential solutions. Applying NLP techniques to data from X may offer insights into self-reported posts, medication use, and medication-related side effects across breast cancer patients with various demographics, potentially surpassing the depth and breadth of traditional cohort studies. As such, the current study has been designed to:


\begin{itemize}
\item Identify self-disclosures of breast cancer from social media, build a cohort, and collect longitudinal data.

\item Conduct NLP-driven analyses to detect and uncover patterns of medication usage among breast cancer patients and medication-associated side effects.

\item Generate detailed statistics associated with the distribution of side effects observed across breast cancer-approved medications and identify potentially unknown medication-side effect associations.
\end{itemize}

\section{Materials and Methods}

\subsection{Dataset}
We collected a large number of breast cancer-related posts (n = 1,454,637) from X. This dataset was compiled using four specific keywords: `\textit{cancer}', `\textit{breastcancer}' (as a single term),  `\textit{tamoxifen}', and `\textit{survivor}', along with their hashtag equivalents (e.g., `\textit{\#breastcancer}'). An analysis of data collected using specific keywords revealed that although there were numerous health-related posts from genuine breast cancer patients, they were accompanied and often obscured by a significant amount of content posted by people who were not breast cancer patients/survivors (e.g., people sharing awareness about breast cancer). Four annotators in a study conducted by Al-Garadi et al. \cite{al2020automatic} processed and labeled a subset of 5,019 unique posts of this dataset into two classes: a) self, a family member, or friend-report of breast cancer (S), and b) not relevant posts (NR). The intuition behind this annotation was that if subscribers on X self-disclosed breast cancer statuses, those disclosures could be leveraged to create a social media based breast cancer cohort. We have used this annotated data with the same train-test split for the supervised model development to extract the relevant posts from the full dataset for breast cancer medication and associated side effects analysis.

\subsection{Self-reported breast cancer post, medication, and side effect discovery}
We adopt three distinct approaches to tackling the supervised classification of social media posts. Firstly, we extracted various feature sets from the text and constructed eight different classical machine learning classifiers. For this aim, we explored the combination of a broad spectrum of features, including n-grams (ranging from 1 to 3), word clusters \cite{sarker2017corpus}, word-to-vector representations, text length, term frequency-inverse document frequency (TF-IDF), latent Dirichlet allocation (LDA) (i.e., extracted features based on latent topics of a post), sentiment score, and bidirectional encoder representations from transformers (BERT) embeddings, as the input features for developing different machine learning models. These models were then optimized through a grid search method, involving an extensive range of parameters and 5-fold cross-validation on the training dataset. Additionally, we developed a two-layer BLSTM model (parameters: unit = 100, dropout = 0.2, recurrent dropout = 0.2), followed by a dense layer (parameters: unit = 100, dropout = 0.2). We also fine-tuned the pre-trained transformer-based architectures and weights of the BERT and BERT large models on the available training dataset.

\par We created two lexicons manually from the annotated text to represent medication expressions and their associated side effects. For this aim, the lexicons were built upon the available National Cancer Institute medication library\footnote{\url{https://www.cancer.gov/about-cancer/treatment/drugs/breast}. [Accessed 02-19-2024]} and side effects\footnote{\url{https://www.cancer.gov/about-cancer/treatment/side-effects}. [Accessed 02-19-2024]} as well as COVID-19 symptoms \cite{sarker2020self} lexicons. We specifically explored medications that have been approved by the Food and Drug Administration (FDA) for breast cancer treatment to capture the associated side effects.

\par
\begin{figure*}[h]
\centering
    \includegraphics[width=1\textwidth]{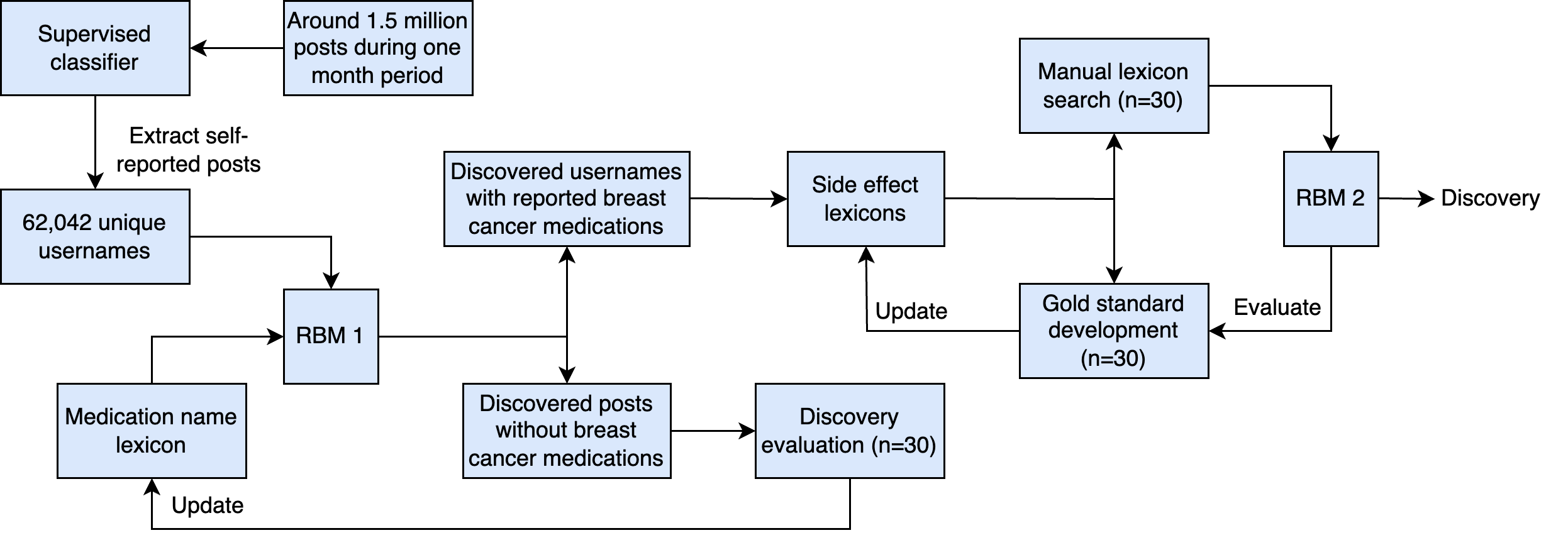}
    \caption{Flow diagram of the methods for medication and their associated side effects discovery from the social media cohort data. Abbreviations: RBM, rule-based model.}
    \label{fig:flow}
\end{figure*}

 The posts were collapsed by unique usernames as we observed that the time span of the available posts was one month, by which we assumed that breast cancer medication prescription patterns and side effects are constant in this particular time span. Then, multiple independent annotators manually annotated these collapsed posts, and the newly found medications and side effects were added to the lexicons. To assess and monitor the performance of the rule-based models over the enrichment of the lexicons, we considered random annotated usernames as the gold standard set. Figure \ref{fig:flow} illustrates the designed workflow for the multi-layer rule-based model development to extract breast cancer medications and their associated side effects. The multi-layer rule-based model is composed of two distinct models, each tasked with identifying medications and side effects separately, employing the concept of inexact matching. Notably, Levenshtein string similarity was used for recognition, accommodating near-misspellings and paraphrased expressions. The models utilized a rolling sliding window, ranging from 1 to 9, with a stride of one, to capture both single-word and multi-word entities. Of note, the model was engineered to prevent the redundant detection of the same words using different window sizes. They also incorporated a feature for negation detection using a list of negation triggers. If a negation is detected, it is flagged accordingly in the final result. Our approach was particularly focused on precision (at the expense of recall) based on the fact that there is no shortage of data from a social media based cohort, and, consequently, avoiding false positives is more important than avoiding false negatives.

\subsection{Descriptive analysis of medications and side effects}
We first defined a breast cancer cohort using the best-performing classifier and applied our rule-based models to identify mentions of breast cancer medications and side effects. To describe the distribution of side effects, we classified breast cancer-approved medications using biological mechanism-based functional classification, specifically hormone therapy, chemotherapy, immune checkpoint inhibitors, and kinase inhibitors. Recognizing that a single medication could be mentioned multiple times in aggregated posts based on unique members in the cohort, we considered only unique occurrences of medications or side effects per sample. As cancer treatment regimen in general may consist of several medications, several functional classifications of medication appeared for each cohort member. Therefore, we first identified the patterns of medications and then tested the association between these medication patterns and side effects using the Kruskal-Wallis test. Multiple tests across side effects were adjusted using Benjamini Hochberg correction. Pair-wise comparisons across medication patterns were performed using a Dunn's test. 

The scripts used in this study were implemented using Python (version: 3.8.8) and R (version: 4.3.2). The level of significance was set to 0.05.

\section{Results}
\subsection{Self-reported breast cancer post, medication, and side effect discovery}
Of the developed supervised classifier, the transformer-based BERT-large language model achieved the highest performance with an accuracy of 0.93 and an $F_{1}$ score of 0.89 on the test data. As such, we used this model as the supervised classifier in the proposed workflow. The model was trained with a maximum sequence length of 100 and batch size of 16 during 40 epochs. During the rule-based model development, the manual annotation was done by three independent annotators, and the pair-wise inter-annotator agreements were calculated using the Cohen's Kappa measure \cite{viera2005understanding}. In the initial annotation round, the average agreement was 0.78. This average increased to 1.0 in the second round, following discussions among the annotators. The multi-layer rule-base model successfully identified breast cancer-approved medications and associated side effects with an $F_{1}$ score of 0.64, precision of 0.64, and recall of 0.64.

\subsection{Descriptive analysis of medications and side effects}
\begin{figure}[h!]
\centering
    \includegraphics[width=1\textwidth, height = 0.5\textwidth]{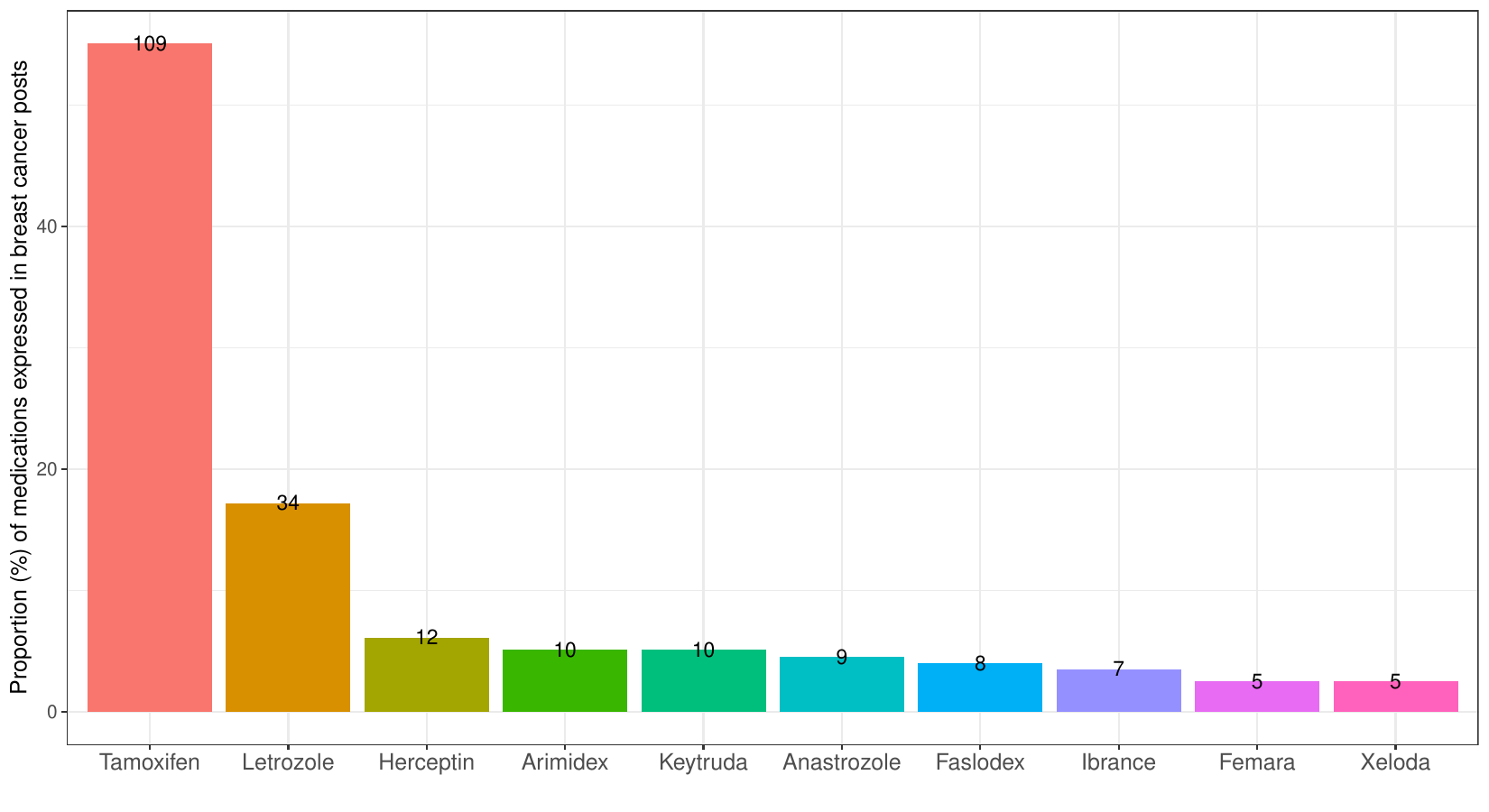}
    \caption{Top 10 most expressed breast cancer approved medications in our social media cohort. The label on top of the bar charts represents the number of cohort members who expressed medications.}
    \label{fig:expressed_drug}
\end{figure}

\begin{figure}[h!]
\centering
    \includegraphics[width=1\textwidth, height = 0.8\textwidth]{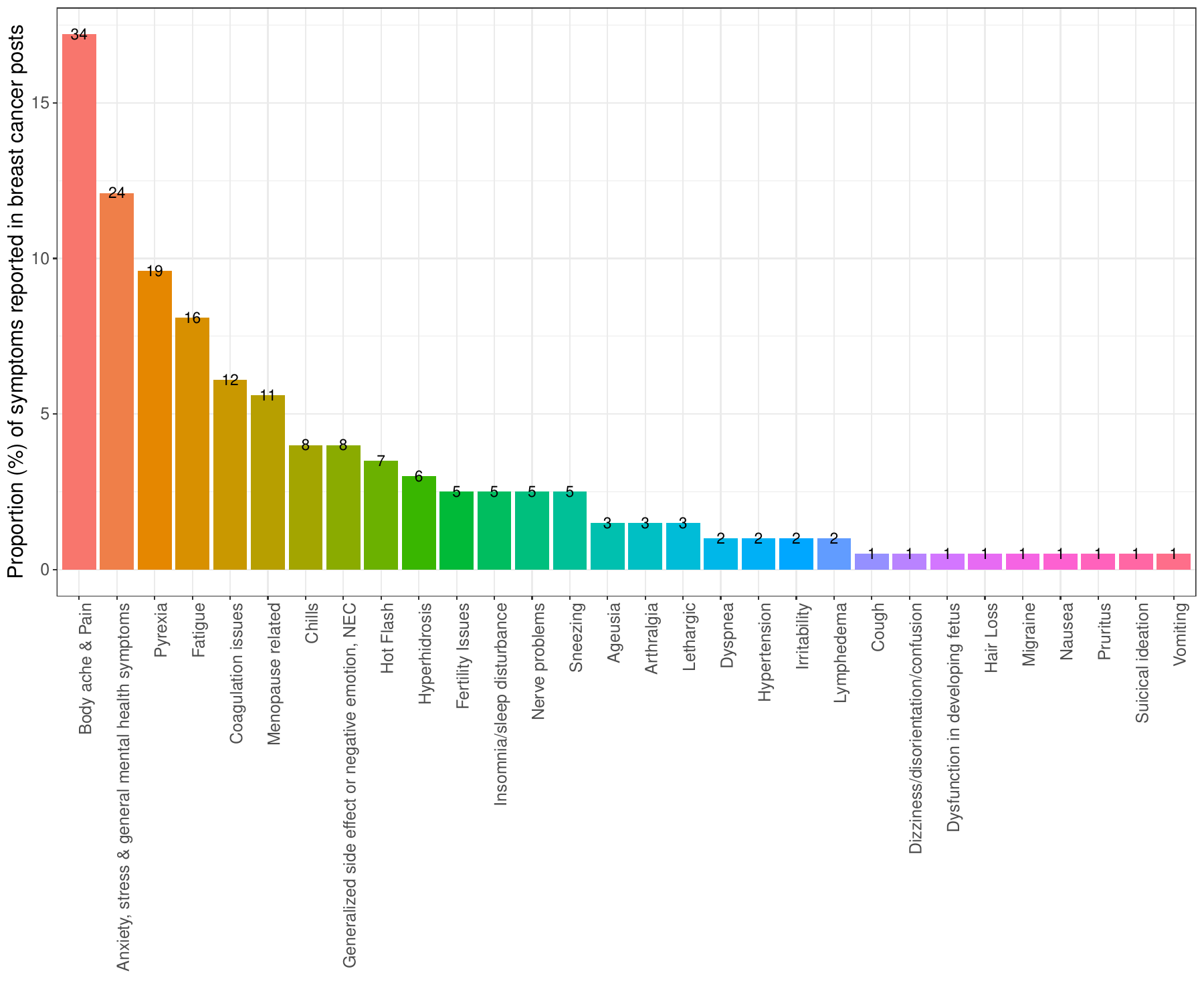}
    \caption{Expressed side effects in our social medial cohort. The y-axis represents the proportions and the text labels on top of bar charts are the number of users who expressed side effects. NEC, not elsewhere classified.}
    \label{fig:side_effect_expressed}
\end{figure}

\begin{figure}[h!]
\centering
    \includegraphics[width=0.8\textwidth, height = 0.6\textwidth]{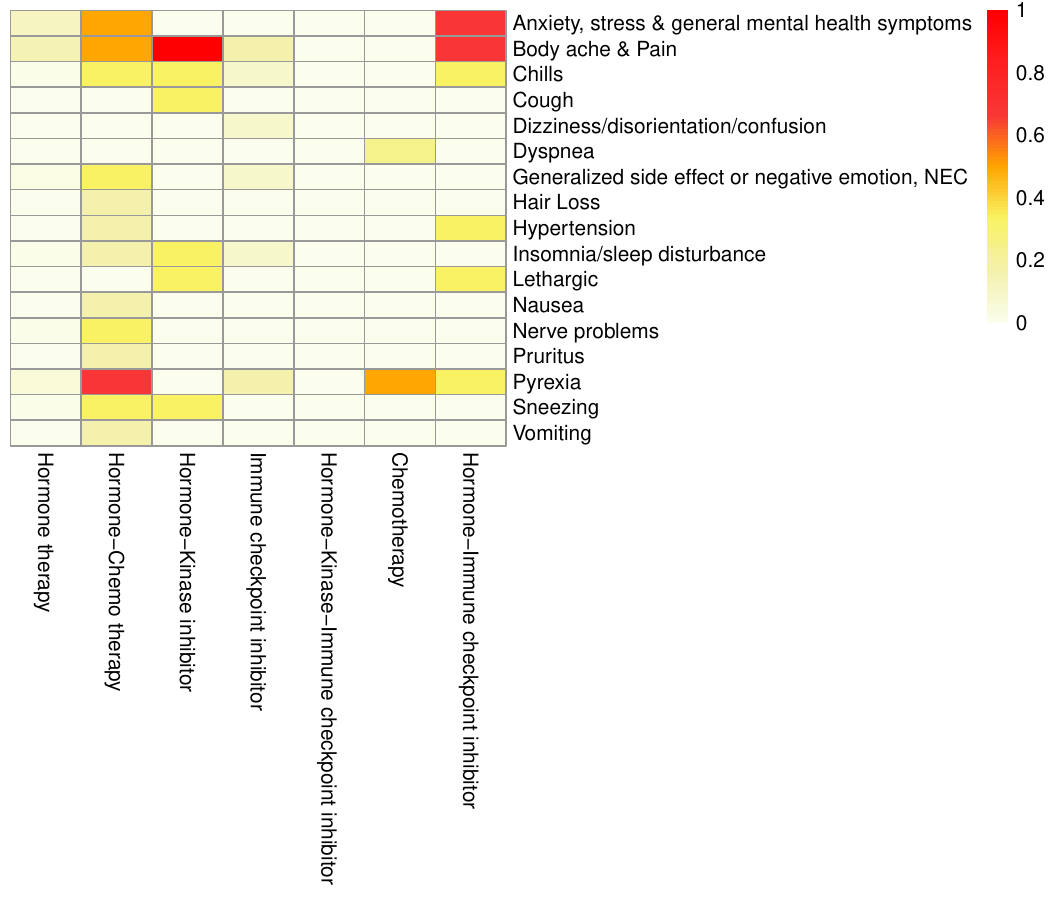}
    \caption{The heat map of prevalence of significantly associated side effects with medication patterns (adjusted p-value $<$ 0.05). NEC, not elsewhere classified.}
    \label{fig:sig_side_effect}
\end{figure}

We detected multiple medication mentions in the discovered breast cancer posts collapsed by unique usernames. Among the final cohort (n = 62,042), at least 198 members expressed a minimum of one breast cancer-approved medication. Many cohort members mentioned taking medications without specifying their names, and we excluded all such cases. 109 (55.1\%) mentioned tamoxifen, and hormone therapy was the most expressed medication category in the posts. Figure \ref{fig:expressed_drug} presents the full distribution of medication mentions. 31 side effects were identified, and the most commonly expressed side effect was body ache \& pain (34 [17.2\%]) (Figure \ref{fig:side_effect_expressed}). In the discovered breast cancer cohort, seven patterns of medication patterns were identified including hormone therapy, chemotherapy, a combination of hormone therapy and chemotherapy, a combination of hormone therapy and kinase inhibitor, immune checkpoint inhibitor, a combination of hormone therapy, kinase inhibitor, and immune checkpoint inhibitor, and a combination of hormone therapy and immune checkpoint inhibitor. A Kruskal-Wallis test showed that 17 out of 31 side effects were significantly associated with the medication patterns (adjusted p-value $<$0.05), of which known side effects of hormone or chemotherapy include pyrexia, body ache, anxiety, nerve problems, and hair loss. In addition, our novel breast cancer-associated side effect lexicon discovered generalized side effect or negative emotion, not elsewhere classified (NEC), which represents a subjective feeling towards cancer and medications, such as `\textit{worst feeling'} or `\textit{feeling of dreadful side effect'} (Figure \ref{fig:sig_side_effect}). The prevalence of generalized side effect or negative emotion, NEC in a combination of hormone therapy and chemotherapy was significantly higher than the prevalence in hormone therapy (adjusted p-value $<$0.05).

\section{Discussion}
We trained a transformer-based model to identify self-reported breast cancer posts on social media and create a cohort. The developed supervised transformer-based classifier demonstrates superior performance compared to the classical machine learning methods that worked with the extracted features, thanks to its innovative architecture and pre-training scheme. This finding highlights the feasibility of constructing a substantial breast cancer cohort from social using an automated NLP pipeline and detecting breast cancer therapy-associated side effects using lexicon development. The methods may also be replicated to create other similar cohorts. 

Using a multi-layer rule-based model architecture that was optimized for precision, we detected medication name expressions and side effects in each unique cohort member profile. In our analysis of accounts discussing breast cancer, hormone therapy was the most expressed medication category, with tamoxifen being the most commonly mentioned keyword. By developing a novel breast cancer therapy-associated side effect lexicon, we identified patterns of side effects that were related to medication patterns. 

We discovered that breast cancer therapy is associated with a broad range of side effects, as expressed by the cohort members. Chemotherapy is associated with a number of neurological side effects, including nausea, pain, and hair loss \cite{pearce2017incidence}. Kinase inhibitors for breast cancer are associated with adverse and side effects in the cardiovascular system, such as hypertension, atrial fibrillation, and heart failure, gastrointestinal, and skin reactions \cite{shyam2023adverse,iancu2022tyrosine,le2021role}. Clinical trials of immune checkpoint inhibitors indicated that colitis and pneumonitis are the most frequent fatal adverse effects of immune checkpoint inhibitors \cite{wang2018fatal}. Notably, we were able to identify well-known side effects of hormone therapy and chemotherapy, such as pyrexia, body ache, anxiety, and nerve problems. Furthermore, our novel breast cancer therapy-associated side effect lexicon discovered generalized side effect or emotion, NEC. Although this lacks a detailed description of side effects, our cohort users may suffer from an indescribable feeling, which may be a pre-clinical side effect of emotional distress or common side effects of breast cancer therapy.

Our work demonstrates the utility of a social media-based cohort that is created automatically via NLP and machine learning methods for identifying patterns of medications and side effects. Such cohorts, once the methods are established and deployed, can grow automatically over time, leading to the collection of seemingly unlimited data. Potentially novel insights may then be mined using the strategies we described in this paper. The discovered side effects using our novel breast cancer-associated side effect lexicon represent potential hypotheses that can be studied and validated through more traditional studies. Such methods of cohort data analyses may be particularly useful for new medications entering the market for which post-marketing surveillance data is limited or absent. Such strategies may also enable the early detection of potential unknown side effects. Also, while the side effects discussed on social media may not be severe (e.g., nausea), they may be the reasons for non-adherence among patients, an association that needs to be investigated in future research.



\par Despite the strength of our analytical framework, several aspects should be considered for future improvement in the analysis. First, social media posts often lacked enough context for accurate classification, leading to potential mis-identification of breast cancer-related posts. Second, we collapsed the posts based on accounts, assuming the homogeneity of the medications' prescription regimens and associated side effects. Multiple medication expressions appearing in one of the cohort member's profiles can lead to false positive indications of the association between medication and its side effects. Third, of around 1.5 million available posts, only 198 cohort members were detected to specify the name of at least one breast cancer-approved medication name (non-standard or generic expressions were potentially missed in our analyses). This small sample size may underpower our lexicon to discover a side effect occurrence. 

\section{Conclusion}
A supervised classifier was able to identify a self-reported breast cancer cohort. Multiple rounds of lexicon development of medications and side effects were conducted, and rule-based models were designed to describe the medication usage prevalence and their links to side effects. We demonstrate, for the first time, the feasibility of an NLP model discovering the patterns of side effects associated with breast cancer-approved medications. Notably, our breast cancer therapy-associated side effect lexicon identified a potential pre-clinical side effect in breast cancer therapy. The next steps include investigating the proposed workflow in a larger sample size and other social media platforms. This can involve the consideration of the usage of non-breast cancer medications and assessing the magnitude of side effects alleviation due to supportive medications.

%
%
%
\bibliographystyle{splncs04}
\bibliography{ref}
%




\end{document}